# Определение позиций резки сотоблоков с использованием технологий компьютерного зрения

**А.В. Разумовский, Я.Ю. Пикалов, М.В.Сарамуд**

Сибирский государственный университет науки и технологий имени академика М. Ф. Решетнева, Российская Федерация, 660037, г. Красноярск, просп. им. газ. «Красноярский рабочий», 31

**Аннотация**

В статье рассматривается способ автоматизации процесса резки сотоблока, а конкретно получение точек и углов резки требуемых граней. При расчетах учитываются следующие требования: допустимое расположение плоскости реза - 0,4 от длины грани ячейки, плоскость реза должна быть перпендикулярна стенке ячейки. Сам алгоритм состоит из двух основных этапов: определение структуры сотоблока и поиск точек реза. При отсутствии существенных дефектов у сотоблоков (деформация профиля ячейки и вмятина на грани ячеек) алгоритм определения структуры работает без существенных неточностей. Результаты работы алгоритма поиска точек реза также удовлетворительны.

**Введение**

В процессе конструирования космических и летательных аппаратов задействуются различные компоненты с низкой массой и высокой прочностью. Одними из подобного рода компонентов являются сотопанели [1-3], состоящие из двух листов углепластиковых или алюминиевых листов, приклеенных к сотовому наполнителя с гексагональными ячейками (рисунок 1).

На данный момент среди основных методов обработки данного рода изделий выделяется фрезерование и разрезание стенок сот специальным ножом. Последний метод является наиболее распространенным в связи с возможностью получения наиболее сложных пространственных форм сотоблоков [4].

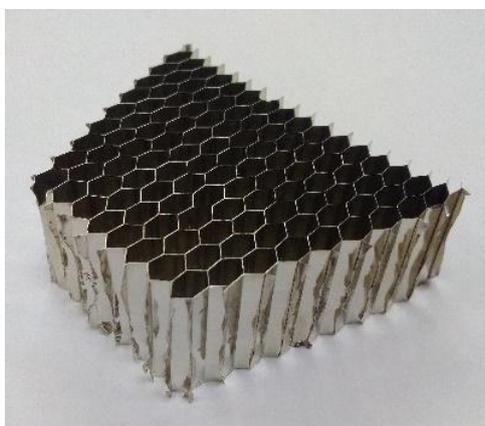

Рисунок 1 – Ячеистая структура сотоблока

Автоматизация процесса обработки сотоблоков путем их резки подразумевает анализ их структуры и непосредственно определения точек реза. При этом, согласно проведенным исследованиям [5], необходимо учитывать допустимое расположение плоскости реза, равное 0,4 от длины стенки ячейки. Также плоскость реза должна быть перпендикулярна стенки ячейки. Соблюдение данных условий позволяет избежать деформации соседних стенок ячеек.

В качестве исходных данных для анализа и вычислений предполагается использовать фотографии изделия, полученные с видеокамеры, а также конечную форму изделия, которая задается как один из параметров обработки. Рассмотрим этапы всего алгоритма отдельно.

**Определение структуры сотоблока**

С использованием нейронной сети, но обученной под данную задачу, производится поиск всех точек, соответствующих узловым точкам ячеек на изображении сотоблока. Использование нейронной сети позволяет адаптировать алгоритм к различным условиям, в которых планируется производить обработку (освещение, фон и т.д.). После этого на полученном бинарном изображении отбираются точки, яркость которых превышает 0.3 для того, чтобы исключить ложные определения, затем определяются координаты оставшихся точек.

Общий порядок действий данного этапа можно представить в следующих шагах:
1. Использование нейронной сети для поиска узловых точек.
2. Отбор точек, яркость которых превышает 0.3 для отсеивания ложных данных.
3. Определение координат отобранных точек.
4. Определение граней
    a. Определение среднего расстояния от каждой узловой точки до 3 ближайших других.
    b. Определение для каждой узловой точки соседних узловых точек. Соседними узловыми точками (концы отрезков, характеризующих грани) считаются точки, расстояние между которыми находится в диапазоне [0.5; 1.3] от среднего определенного ранее расстояния.

На рисунках 2 и 3 представлены примеры результатов работы данного этапа.

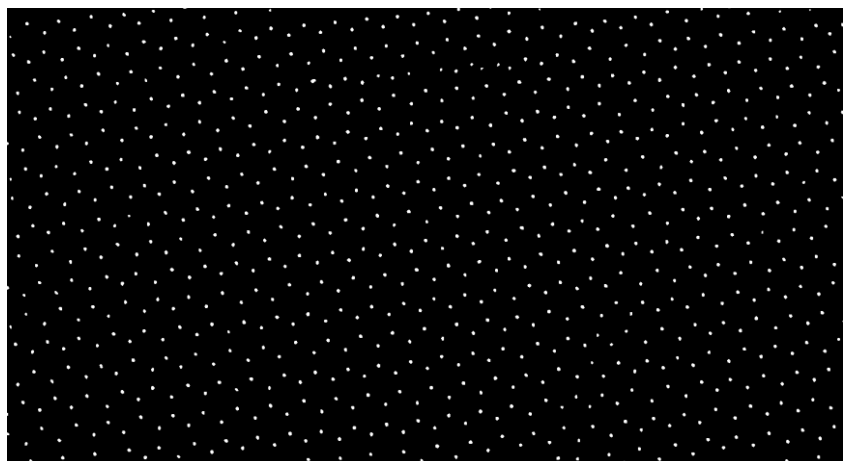

Рисунок 2 – Определенные на шаге 1 узловые точки

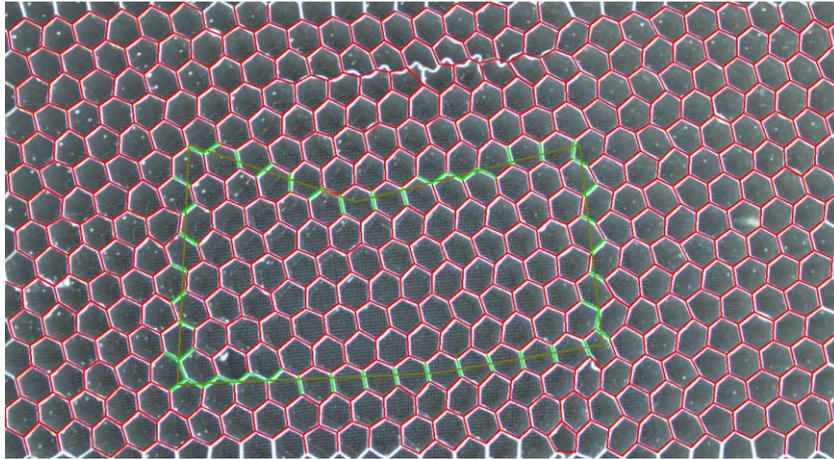

Рисунок 3 – Обозначения: красные и зеленые отрезки – определенные в ходе работы алгоритма грани,
темно-зеленый контур – форма заданной для вырезки фигуры.

**Определение точек реза**

Данный этап является заключительным, где непосредственно осуществляется поиск и формирование списка точек реза. Последовательность шагов:

1. Построение маски по контуру, соответствующей форме заданного изделия (рисунок 4.а).
2. Построение маски, соответствующей форме заданного изделия с учетом структуры сотоблока.
3. Модификация маски, полученной на шаге 2 путем расширения области на значение, равное произведению средней длины грани сотоблока на коэффициент, характеризующий долю грани от края заготовки, недопустимую для резки (0,4) (рисунок 4.б).
4. Определение контура области, полученной на шаге 3.
5. Определение точек пересечения граней и контура, полученного на шаге 4.
6. Определение угла резки с использованием данных о местоположении грани, на которой находится данная точка. Угол разреза относительно грани должен равняться 90 градусам.
7. Упорядочивание найденных точек.
    a. Выбирается некоторая начальная точка.
    b. Определяется ближайшая к ней точка, которая будет считаться следующей.
    c. Далее для определенной на шаге b точки шаг b повторяется уже для неё.

На рисунках 4 и 5 представлены примеры работы данного этапа.

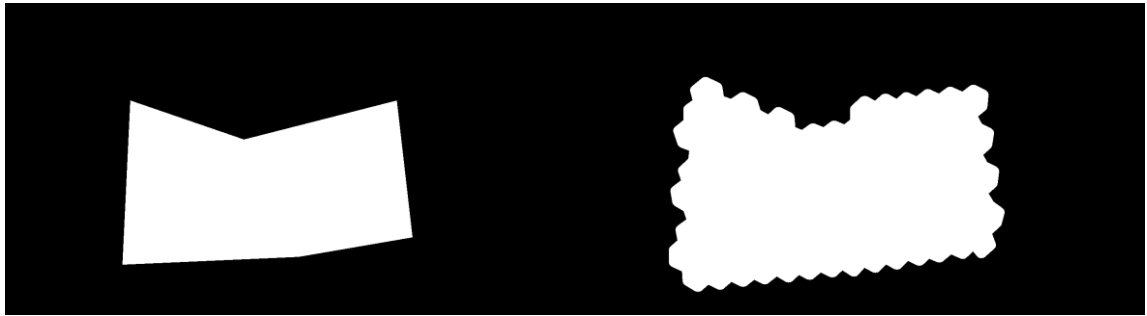

(а) (б)

Рисунок 4 – Маски, полученные на шагах 1-3: а) маска по контуру изделия, б) модифицированная маска

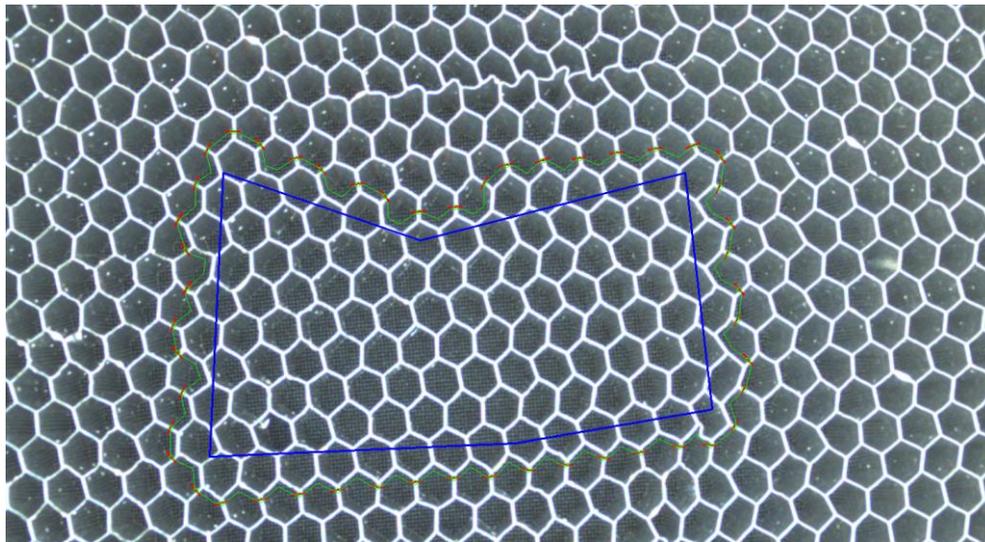

Рисунок 5 – Обозначения: синий контур – исходная форма изделия; зеленый контур – конечная форма изделия; красные отрезки – места и углы резки сотоблока

В завершение производится сортировка порядка точек в полученном списке для уменьшения времени, затрачиваемого роботом-манипулятором на перемещение ножа.

**Вывод**

Рассмотрен подход определения структуры сотоблока с использованием обученной U-Net-подобной нейронной сети с последующей алгоритмической обработкой. Получаемые при работе данного алгоритма результаты на фотографиях сотоблоков без дефектов не содержат существенных неточностей.

Также реализован алгоритм поиска точек реза для вырезки изделия с сохранением обязательных требований: угол резки относительно положения грани 90 градусов, отступ резки от узловой точки не менее 0,4 длины всей грани. Результаты работы данного этапа всего алгоритма также не имеют существенных неточностей.

Среди допустимых улучшений данного алгоритма рассматривается вариант модификации этапа определения структуры сотоблока путем задействования методов и алгоритмов, позволяющих выделить грани сотоблока и получить "скелет" изделия.

**Список литературы**

[1] V. Crupi, G. Epasto, E. Guglielmino, "Comparison of aluminium sandwiches for lightweight ship structures: Honeycomb vs. foam," *Marine Structures*, vol. 30, 2013, pp. 74-96, https://doi.org/10.1016/j.marstruc.2012.11.002.

[2] A. Boudjemai, M. H. Bouanane, A. Mankour, H. Salem, R. Hocine and R. Amri, "Thermo-mechanical design of honeycomb panel with fully-potted inserts used for spacecraft design", *2013 6th International Conference on Recent Advances in Space Technologies (RAST)*, Istanbul, 2013, pp. 39-46, doi: 10.1109/RAST.2013.6581238.

[3] A. Boudjemai, M. H. Bouanane, A. Mankour, H. Salem, R. Hocine and R. Amri, "Thermo-mechanical design of honeycomb panel with fully-potted inserts used for spacecraft design", *2013 6th International Conference on Recent Advances in Space Technologies (RAST)*, Istanbul, 2013, pp. 39-46, doi: 10.1109/RAST.2013.6581238.

[4] G. Bianchi, G. S. Aglietti and G. Richardson, "Development of efficient and cost-effective spacecraft structures based on honeycomb panel assemblies," *2010 IEEE Aerospace Conference*, Big Sky, MT, 2010, pp. 1-10, doi: 10.1109/AERO.2010.5446748.

[5] Application of sequential processing of computer vision methods for solving the problem of detecting the edges of a honeycomb block / M. V. Kubrikov, I. A. Paulin, M. V. Saramud, A. S. Kubrikova // Journal of Physics: Conference Series, Krasnoyarsk, Russian Federation, 25 сентября – 04 2020 года. – Krasnoyarsk, Russian Federation, 2020. – P. 42098. – DOI 10.1088/1742-6596/1679/4/042098.